\documentclass[conference,draft,11pt,onecolumn]{IEEEtran}

\usepackage{cite}
\usepackage{amsmath,amssymb}
\usepackage{qtree}
\usepackage{float}

\newcommand{\Tab}[1]{Tab. \ref{#1}}
\newcommand{\pref}[1]{(\ref{#1})}
\newcommand{\TT}[1]{\texttt{#1}}
\def\seq#1{\mathbf{#1}}

\begin{document}

\title{\bf Reinforcement Learning of \\
    Minimalist Numeral Grammars}

\author{\IEEEauthorblockN{
    Peter beim Graben\IEEEauthorrefmark{1}, Ronald R\"{o}mer\IEEEauthorrefmark{1},
    Werner Meyer\IEEEauthorrefmark{1}, Markus Huber\IEEEauthorrefmark{1},
    and Matthias Wolff\IEEEauthorrefmark{1}
    }
\IEEEauthorblockA{\IEEEauthorrefmark{1}Brandenburgische Technische Universit\"{a}t Cottbus -- Senftenberg, \\
    Institute of Electronics and Information Technology, \\
    Department of Communications Engineering, \\
    Cottbus, Germany}
 peter.beimgraben@b-tu.de
}

\maketitle

\begin{abstract}
Speech-controlled user interfaces facilitate the operation of devices and household functions to laymen.
State-of-the-art language technology scans the acoustically analyzed speech signal for relevant keywords that are subsequently inserted into semantic slots to interpret the user's intent. In order to develop proper cognitive information and communication technologies, simple slot-filling should be replaced by utterance meaning transducers (UMT) that are based on semantic parsers and a \emph{mental lexicon}, comprising syntactic, phonetic and semantic features of the language under consideration. This lexicon must be acquired by a cognitive agent during interaction with its users. We outline a reinforcement learning algorithm for the acquisition of the syntactic morphology and arithmetic semantics of English numerals, based on minimalist grammar (MG), a recent computational implementation of generative linguistics. Number words are presented to the agent by a teacher in form of utterance meaning pairs (UMP) where the meanings are encoded as arithmetic terms from a suitable term algebra. Since MG encodes universal linguistic competence through inference rules, thereby separating innate linguistic knowledge from the contingently acquired lexicon, our approach unifies generative grammar and reinforcement learning, hence potentially resolving the still pending Chomsky-Skinner controversy.
\end{abstract}

\IEEEpeerreviewmaketitle



\section{Introduction}
\label{sec:intro}

Speech-controlled user interfaces such as Amazon's \emph{Alexa}, Apple's \emph{Siri} or \emph{Cortana} by Microsoft substantially facilitate the operation of devices and household functions to laymen. Instead of using keyboard and display as input-output interfaces, the operator pronounces requests or instructions to the device and listens to its responses.

State-of-the-art language technology scans the acoustically analyzed speech signal for relevant keywords that are subsequently inserted into semantic frames \cite{Minsky74} to interpret the user's intent. This \emph{slot filling} procedure \cite{Allen03, TurHakkaniEA11, MesnilDauphinEA15} is based on large language corpora that are evaluated by standard machine learning methods, such as conditional random fields \cite{TurHakkaniEA11} or by deep learning of neural networks \cite{MesnilDauphinEA15}, for instance. The necessity to overcome traditional slot filling techniques by proper cognitive information and communication technologies has already been emphasized by Allan \cite{Allen17}. His research group trains semantic parsers from large language data bases such as WordNet or VerbNet that are constrained by hand-crafted expert knowledge and semantic ontologies \cite{Allen03, Allen14, AllenBahkshandehEA18}.

One particular demand on cognitive user interfaces are the processing and understanding of numerals, e.g. in instructions like ``\texttt{increase the heating to 22.5 degrees}'', where the device may probably respond with a sensor registration: ``\texttt{the current room temperature is 18.3 degrees}'' \cite{TschopeDuckhornEA18}. Numerals are an important research domain in cognitive linguistics and language technology \cite{FlachHolzapfelEA00, WolffEichnerHoffmann01, Hurford01, IoninMatushansky06, Mendia18, GrabenMeyerEA19}. They exhibit typological differences among languages but share a simple arithmetic semantics. Decent examples are different morphologies in German ($\TT{zweiundvierzig} = 2 + 40$) or English ($\TT{fourtytwo} = 40;2$), and also different base systems in German ($\TT{achtzig} = 8 \times 10$) or French ($\TT{quatre\text{-}vingts} = 4 \times 20$) \cite{Hurford01}. Linguistically, numerals are regarded as modifiers \cite{IoninMatushansky06} with a particular syntactic morphology that should be described by a suitable grammar formalism. This grammar must store numeral morphemes together with their arithmetic semantics in a data base, called the \emph{mental lexicon}. It should be complex enough to account for the wealth of linguistic typology and constrained enough to exclude ungrammatical compositions such as $\TT{zweizig}$ in German or $\TT{twoty}$ in English \cite{Hurford01}.

Recent research in computational linguistics has demonstrated that quite different grammar formalisms, such as categorial grammar \cite{AmblardLecomteRetore10}, tree-adjoining grammar \cite{JoshiLevyTakahashi75}, multiple context free grammar (MCFG) \cite{SekiEA91}, range concatenation grammar \cite{Boullier05}, and minimalist grammar \cite{Stabler97, StablerKeenan03} converge toward universal description models \cite{Michaelis01a, Stabler11a}. Minimalist grammar has been developed by Stabler \cite{Stabler97} to mathematically codify Chomsky's \emph{Minimalist Program} \cite{Chomsky95} in the generative grammar framework. A minimalist grammar (MG) consists of a mental lexicon storing linguistic signs as arrays of syntactic, phonetic and semantic features, on the one hand, and of two structure-building functions, called ``merge'' and ``move'', on the other hand. Syntactic features in the lexicon are, e.g., the linguistic base categories noun ($\TT{n}$), verb  ($\TT{v}$), adjective ($\TT{a}$), or, in the present context, numeral ($\TT{num}$). These are syntactic heads selecting other categories either as complements or as adjuncts. The structure generation is controlled by selector categories that are ``merged'' together with their selected counterparts. Moreover, one distinguishes between licensors and licensees, triggering the movement of maximal projections. An MG does not comprise any phrase structure rules; all syntactic information is encoded in the feature array of the mental lexicon. Furthermore, syntax and compositional semantics can be combined via the lambda calculus \cite{Niyogi01, Kobele09}, while MG parsing can be implemented by compilation into an equivalent MCFG \cite{Stabler11b}.

One important property of MG is their effective learnability in the sense of Gold's formal learning theory \cite{Gold67}. Specifically, MG can be acquired by positive examples \cite{KobeleCollierEA02, StablerEA03} from linguistic dependence graphs \cite{BostonHaleKuhlmann10, KleinManning04}, which is consistent with psycholinguistic findings on early-child language acquisition \cite{Ellis06, Pinker95, Tomasello06}. However, learning through positive examples only, could easily lead to overgeneralization. According to Pinker \cite{Pinker95} this could effectively be avoided through reinforcement learning \cite{Skinner15, SuttonBarto18}. Although there is only little psycholinguistic evidence for reinforcement learning in human language acquisition \cite{Moerk83, SundbergEA96}, we outline a machine learning algorithm for the acquisition of an MG mental lexicon of numeral morphology and semantics through reinforcement learning in this contribution.


\section{Numeral Grammar}
\label{sec:ng}

Our language acquisition approach for numeral grammar combines methods from computational linguistics, formal logic, and abstract algebra. Starting point of our algorithm are \emph{utterance meaning pairs} (UMP)
\begin{equation}\label{eq:ump}
    u = \langle e, \sigma \rangle \:,
\end{equation}
where $e \in E$ is the spoken or written utterance, given as the \emph{exponent} of a linguistic sign \cite{Kracht03}. Technically, exponents are strings taken from the Kleene hull of some finite alphabet, $A$, i.e. $E = A^*$. The sign's \emph{semantics} $\sigma \in \Sigma$ is a logical term, usually expressed by means of lambda calculus.


\subsection{Numeral Semantics}
\label{sec:nsem}

The straightforward meaning of a numeral, say $\TT{fourtytwo}$, is  a number concept, such as $42$. However, from a computational point of view, the UMP $\langle \TT{fourtytwo}, 42 \rangle$ simply relates a symbolic string $\TT{fourtytwo}$ to another symbolic string $42$, without making the exponent and the semantics of the sign operationally accessible. This is achieved by interpreting digit strings in a $g$-adic number system. In the decimal system with $g = 10$, we have
\begin{equation}\label{eq:42}
  42 = 4 \times 10 + 2 \times 1 = \sum_{k = 1}^n a_k g^{k-1}
\end{equation}
with $n$ coefficients $0 \le a_k \le g - 1$ ($n$ the number of digits).

Equation \pref{eq:42} can directly be written as a tree-like arithmetic term structure
\begin{figure}[H]
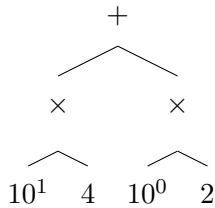

\Tree [.$+$
        [.$\times$
                $10^1$ $4$
            ]
            [.$\times$
                   $10^0$ $2$
            ]
]
\caption{\label{fig:mwr42} Arithmetic term tree for $42$.}
\end{figure}

Using the binary operators $+(x, y) = x + y$ and $\times(x, y) = x \times y$, and writing them in the unary Sch\"onfinkel representation
\begin{eqnarray*}
  +(x, y) &=& +(y)(x) = x + y \\
  \times(x, y) &=& \times(y)(x) = x \times y
\end{eqnarray*}
where $+(y)$ is regarded as a function $f: x \mapsto (+(y))(x) = x + y =f(x)$, and $\times(y)$ as another function $g: x \mapsto (\times(y))(x) = x \times y = g(x)$, respectively, we obtain an expression of the arithmetic term algebra \cite{Kracht03} in Polish notation
\[
  \sigma = +( \times(10^1)(4) )( \times(10^0)(2) )
\]
that will be interpreted as the \emph{meaning} of the numeral $\TT{fourtytwo}$ in the sequel \cite{Mendia18, GrabenMeyerEA19}. Hence, the correct UMP for $42$ is
\begin{equation}\label{eq:term42}
    u = \langle \TT{fourtytwo}, +( \times(10^1)(4) )( \times(10^0)(2) ) \rangle \:.
\end{equation}


\subsection{Minimalist Grammar}
\label{sec:mg}

Following Kracht \cite{Kracht03}, we regard a linguistic sign as an ordered triple
\begin{equation}\label{eq:sign}
    z = \langle e , t , \sigma \rangle
\end{equation}
with the same exponent $e \in E$ and semantics $\sigma \in \Sigma$ as in the UMP \pref{eq:ump}. In addition, $t \in T$ is a syntactic \emph{type} that we encode by means of minimalist grammar (MG) in its chain representation \cite{StablerKeenan03}. The type controls the generation of syntactic structure and hence the order of lambda application, analogously to the typed lambda calculus in Montague semantics.

An MG consists of a data base, the mental lexicon, containing signs as arrays of syntactic, phonetic and semantic \emph{features}, and of two structure-generating functions, called ``merge'' and ``move''. Syntactic features are the \emph{basic types} $b \in B$ from a finite set $B$, with $b = \TT{n}, \TT{v}, \TT{a}, \TT{num}$, etc, together with a set of their respective \emph{selectors} $S = \{ \TT{=}b | b \in B \}$ that are unified by the ``merge'' operation. Moreover, one distinguishes between a set of licensers $L_+ = \{ \TT{+}l | l \in L \}$ and another set of their corresponding licensees $L_- = \{ \TT{-}l | l \in L \}$ triggering the ``move'' operation. $L$ is another finite set of movement identifiers. $F = B \cup S \cup L_+ \cup L_-$ is called the feature set. Finally, one has a two-element set $C = \{ \TT{:\!:}, \TT{:} \}$ of categories, where ``$\TT{:\!:}$'' indicates \emph{simple}, \emph{lexical} categories while ``$\TT{:}$'' denotes \emph{complex}, \emph{derived} categories. The ordering of syntactic features is prescribed as regular expressions, i.e. $T = C (S \cup L_+)^* B L_-^*$ is the set of syntactic types \cite{Stabler97, StablerKeenan03}. The set of linguistic signs is then given as $Z = E \times T \times \Sigma$ \cite{Kracht03}.

Let $e_1, e_2 \in E$ be exponents, $\sigma_1, \sigma_2 \in \Sigma$ semantic terms in the lambda calculus, $f \in B \cup L$ one feature identifier, $\seq{t}, \seq{t}_1, \seq{t}_2 \in F^*$ feature strings compatible with the regular types in $T$, $\cdot \in C$ and $\seq{z}, \seq{z}_1, \seq{z}_2 \in Z^*$ sequences of signs, then $\langle e_1 , \TT{:\!:} \TT{=}f \seq{t}_1, \sigma_1 \rangle$ and $\langle e_2 , \TT{:} f , \sigma_2 \rangle$ form signs in the sense of \pref{eq:sign}. A sequence of signs is called a \emph{minimalist expression}, and the first sign of an expression is called its \emph{head}, controlling the structure building through ``merge'' and ``move'' as follows.

The MG function ``merge'' is defined through inference schemata
\begin{eqnarray}
&&\dfrac{
    \langle e_1, \TT{:\!:=}f \seq{t}, \sigma_1 \rangle \quad
    \langle e_2, \cdot f, \sigma_2 \rangle \seq{z}
  }{
    \langle e_1e_2, \TT{:} \seq{t}, \sigma_1\sigma_2 \rangle \seq{z}
  }
  \,\text{merge-1} \:, \label{merge-1} \\
&&\dfrac{
    \langle e_1, \TT{:=}f \seq{t}, \sigma_1 \rangle \seq{z}_1 \quad
    \langle e_2, \cdot f, \sigma_2 \rangle \seq{z}_2
  }{
    \langle e_2e_1, \TT{:} \seq{t}, \sigma_1 \sigma_2 \rangle \seq{z}_1 \seq{z}_2
  }
  \,\text{merge-2} \:, \label{merge-2} \\
&&\dfrac{
    \langle e_1,\cdot \TT{=}f \seq{t}_1, \sigma_1 \rangle \seq{z}_1 \quad
    \langle e_2, \cdot f \seq{t}_2, \sigma_2 \rangle \seq{z}_2
  }{
    \langle e_1, \TT{:} \seq{t}_1, \sigma_1 \rangle \seq{z}_1
    \langle e_2, \TT{:} \seq{t}_2, \sigma_2 \rangle \seq{z}_2
  }
  \,\text{merge-3} \:, \label{merge-3}
\end{eqnarray}
Correspondingly, ``move'' is given through
\begin{eqnarray}
&&\dfrac{
    \langle e_1, \TT{:+}f \seq{t}, \sigma_1 \rangle \seq{z}_1
    \langle e_2, \TT{:-}f, \sigma_2 \rangle \seq{z}_2
  }{
    \langle e_2e_1, \TT{:} \seq{t}, \sigma_1\sigma_2 \rangle \seq{z}_1 \seq{z}_2
  }
  \,\text{move-1} \:, \label{move-1} \\
&&\dfrac{
    \langle e_1, \TT{:+}f \seq{t}_1, \sigma_1 \rangle \seq{z}_1
    \langle e_2, \TT{:-}f \seq{t}_2, \sigma_2 \rangle \seq{z}_2
  }{
    \langle e_1, \TT{:} \seq{t}_1, \sigma_1 \rangle \seq{z}_1
    \langle e_2, \TT{:} \seq{t}_2, \sigma_2 \rangle \seq{z}_2
  }
  \,\text{move-2} \:. \label{move-2}
\end{eqnarray}
where only one sign with licensee $\TT{-}f$ may appear in the expression licensed by $\TT{+}f$ in the head. This so-called \emph{shortest movement constraint} (SMC) guarantees syntactic locality demands \cite{Stabler97, StablerKeenan03}. 

A minimalist derivation terminates when all syntactic features besides only one distinguished \emph{start symbol}, in our case $\TT{num}$, have been consumed. The meaning of rules (\ref{merge-1} -- \ref{move-2}) and their applicability becomes clear in the next section.


\section{Reinforcement Learning}
\label{sec:ml}

The language learner is a cognitive agent $L$ in a state $X_t$, to be identified with $L$'s mental lexicon at training time $t$. At time $t = 0$, $L$ is initialized as a \emph{tabula rasa} with empty lexicon
\begin{equation}\label{eq:lexini}
    X_0 \gets \emptyset
\end{equation}
and exposed to UMPs produced by a continuously counting teacher $T$. The first UMPs given by $T$ are $u_1 = \langle \TT{one}, 1 \rangle$, $u_2 = \langle \TT{two}, 2 \rangle$, $u_3 = \langle \TT{three}, 3 \rangle$, and so forth. Note that we assume $T$ presenting already complete UMPs and not singular utterances to $L$. Thus we avoid the \emph{symbol grounding problem} of firstly assigning meanings $\sigma$ to uttered exponents $e$ \cite{Harnad90}, which will be addressed in future research. Moreover, we assume that $L$ is instructed to reproduce $T$'s counting based on its own numeric understanding. This provides a feedback loop and therefore applicability of reinforcement learning \cite{Skinner15, SuttonBarto18}.

As long as $L$ is not able to detect patterns or common similarities in $T$'s UMPs, it simply adds new entries directly to its mental lexicon, assuming that all numerals have base type $\TT{num}$. Hence, $L$'s state $X_t$ evolves according to the update rule
\begin{equation}\label{eq:dyn1}
    X_t \gets X_{t - 1} \cup \{ \langle e_t, \TT{:\!: num}, \sigma_t \rangle \} \:,
\end{equation}
when $u_t = \langle e_t, \sigma_t \rangle$ is the UMP presented at time $t$ by $T$.

In this way, the mental lexicon $X_{12}$ of simplex numerals in \Tab{tab:mg1} has been acquired at time $t = 12$.

\begin{table}[H]
\caption{\label{tab:mg1} Content of the minimalist lexicon $X_{12}$ of language learner $L$ at time $t = 12$.}
\[
\begin{array}{lll}
    \langle \TT{one}, \TT{:\!: num}, 1 \rangle &
    \langle \TT{two}, \TT{:\!: num}, 2 \rangle &
    \langle \TT{three}, \TT{:\!: num}, 3 \rangle \\
    \langle \TT{four}, \TT{:\!: num}, 4 \rangle &
    \langle \TT{five}, \TT{:\!: num}, 5 \rangle &
    \langle \TT{six}, \TT{:\!: num}, 6 \rangle \\
    \langle \TT{seven}, \TT{:\!: num}, 7 \rangle &
    \langle \TT{eight}, \TT{:\!: num}, 8 \rangle &
    \langle \TT{nine}, \TT{:\!: num}, 9 \rangle \\
    \langle \TT{ten}, \TT{:\!: num}, 10^1 \rangle &
    \langle \TT{eleven}, \TT{:\!: num}, +(\times(10^1)(1))(1) \rangle &
    \langle \TT{twelve}, \TT{:\!: num}, +(\times(10^1)(1))(2) \rangle
\end{array}
\]
\end{table}

The learner is so able to perfectly reproduce the learned entries directly via data base query. As a consequence, the teacher $T$ rewards $L$ thus signalling that it has correctly learned the lexicon $X_{12}$.

When the teacher continues counting: $u_{13} = \langle \TT{thirteen}, +(\times(10^1)(1))(3) \rangle$, $u_{14} = \langle \TT{fourteen}, +(\times(10^1)(1))(4) \rangle$, $u_{15} = \langle \TT{fifteen}, +(\times(10^1)(1))(5) \rangle$ and so on, the learner's pattern matching faculty detects a common affix $\TT{teen}$ in the exponents, and a common function $x \mapsto +(\times(10^1)(1))(x)$ in the semantics of UMPs $u_{13}, u_{14}, \dots u_{19}$.

Thus, in a first step UMP $u_{13}$ is still added to the lexicon according to update rule \pref{eq:dyn1},
\begin{equation}\label{eq:dyn13}
    X_{13} \gets X_{12} \cup \{ \langle \TT{thirteen}, \TT{\!:: num} , +(\times(10^1)(1))(3) \rangle  \} \:.
\end{equation}
However, at time $t = 14$, pattern matching, segmentation and lambda abstraction are performed, leading to a revision \cite{KobeleCollierEA02, StablerEA03}
\begin{eqnarray}
    X_{14} &\gets& X_{13} \setminus \{ \langle \TT{thirteen}, \TT{\!:: num} , +(\times(10^1)(1))(3) \rangle \} \label{eq:dyn14_1}  \\
    X_{14} &\gets& X_{14} \cup \{ \langle \TT{teen}, \TT{: =num} \ \TT{num}, \lambda x.+(\times(10^1)(1))(x) \rangle \} \label{eq:dyn14_2} \\
    X_{14} &\gets& X_{14} \cup \{ \langle \TT{thir}, \TT{:\!: num}, 3 \rangle \} \label{eq:dyn14_3} \:,
\end{eqnarray}
such that in \pref{eq:dyn14_1} the previously learned lexicon $X_{13}$ is revised by removing the entry for the composite $\TT{thirteen}$, followed by adding the complex morpheme $\langle \TT{teen}, \TT{: =num} \ \TT{num}, \lambda x.+(\times(10^1)(1))(x) \rangle$ in \pref{eq:dyn14_2}, and completed in \pref{eq:dyn14_3}. For the morpheme $\langle \TT{four}, \TT{:\!: num}, 4 \rangle$ is already contained in the lexicon, further updating is not required at this time.

Next, $L$ has to correctly reproduce the UMPs $u_{13}$ and $u_{14}$ by invoking its \emph{utterance-meaning transducer} (UMT) \cite{GrabenMeyerEA19}. Consider $u_{13}$, which is now ambiguous with respect to the lexicon entries for $3$. First, $L$ may access data base entries $\langle \TT{thir}, \TT{:\!: num}, 3 \rangle $ and $\langle \TT{teen}, \TT{: =num} \ \TT{num}, \lambda x.+(\times(10^1)(1))(x) \rangle $ and derive the following UMP according to the MG rules (\ref{merge-1} -- \ref{move-2})
\begin{eqnarray}\label{eq:deri13_1}
&&\dfrac{\langle \TT{teen}, \TT{: =num} \ \TT{num}, \lambda x.+(\times(10^1)(1))(x) \rangle
    \qquad
    \langle \TT{thir}, \TT{:\!: num}, 3 \rangle
    }
    {\langle \TT{thirteen}, \TT{: num}, (\lambda x.+(\times(10^1)(1))(x))(3) \rangle
    } \: \text{merge-2} \:. \nonumber \\
\end{eqnarray}
This yields the correct semantics with the lambda calculus
\[
    (\lambda x.+(\times(10^1)(1))(x))(3) = +(\times(10^1)(1))(3) = 13
\]
and the uttered exponent $\TT{thirteen}$, generated by the UMT \cite{GrabenMeyerEA19}, is well-formed and will be rewarded by the teacher.

However, $L$ may alternatively select data base entries $\langle \TT{three}, \TT{:\!: num}, 3 \rangle $ and $\langle \TT{teen}, \TT{: =num} \ \TT{num}, \lambda x.+(\times(10^1)(1))(x) \rangle $ as well. Then
\begin{eqnarray}\label{eq:deri13_2}
&&\dfrac{\langle \TT{teen}, \TT{: =num} \ \TT{num}, \lambda x.+(\times(10^1)(1))(x) \rangle
    \qquad
    \langle \TT{three}, \TT{:\!: num}, 3 \rangle
    }
    {\langle \TT{threeteen}, \TT{: num}, (\lambda x.+(\times(10^1)(1))(x))(3) \rangle
    } \: \text{merge-2} \nonumber \\
\end{eqnarray}
will be derived instead. Although it has the correct semantics $13$, uttering the exponent $\TT{threeteen}$ will be rejected by $T$. Upon the resulting punishment, $L$ has to reconfigure its mental lexicon by introducing additional licenser/licensee pairs, here denoted as $\TT{+k}/\TT{-k}$ \cite{KobeleCollierEA02, StablerEA03}. Table \ref{tab:mg2} displays the result of this reorganization process at some time $n$ later than $t = 19$ when all possible ungrammaticalities have been abandoned.

\begin{table}[H]
\caption{\label{tab:mg2} Content of the minimalist lexicon $X_n$ of language learner $L$ after punishment reorganization at time $n$.}
\[
\begin{array}{lll}
    \langle \TT{one}, \TT{:\!: num}, 1 \rangle &
    \langle \TT{two}, \TT{:\!: num}, 2 \rangle &
    \langle \TT{three}, \TT{:\!: num}, 3 \rangle \\
    \langle \TT{thir}, \TT{:\!: num} \ \TT{-k}, 3 \rangle &
    \langle \TT{four}, \TT{:\!: num} \ \TT{-k}, 4 \rangle &
    \langle \TT{five}, \TT{:\!: num}, 5 \rangle \\
    \langle \TT{fif}, \TT{:\!: num} \ \TT{-k}, 5 \rangle &
    \langle \TT{six}, \TT{:\!: num} \ \TT{-k}, 6 \rangle &
    \langle \TT{seven}, \TT{:\!: num} \ \TT{-k}, 7 \rangle \\
    \langle \TT{eight}, \TT{:\!: num} \ \TT{-k}, 8 \rangle &
    \langle \TT{nine}, \TT{:\!: num}\ \TT{-k}, 9 \rangle &
    \langle \TT{ten}, \TT{:\!: num}, 10^1 \rangle \\
    \langle \TT{eleven}, \TT{:\!: num}, +(\times(10^1)(1))(1) \rangle &
    \langle \TT{twelve}, \TT{:\!: num}, +(\times(10^1)(1))(2) \rangle &
    \langle \TT{teen}, \TT{: =num} \ \TT{+k} \ \TT{num}, \lambda x.+(\times(10^1)(1))(x) \rangle
\end{array}
\]
\end{table}

Now only the data base selection $\langle \TT{thir}, \TT{:\!: num} \ \TT{-k}, 3 \rangle$ and $\langle \TT{teen}, \TT{: =num} \ \TT{+k} \ \TT{num}, \lambda x.+(\times(10^1)(1))(x) \rangle$ leads to a grammatical derivation of the UMT \cite{GrabenMeyerEA19},
\begin{eqnarray}\label{eq:deri13_3}
&&\dfrac{\langle \TT{teen}, \TT{: =num} \ \TT{+k} \ \TT{num}, \lambda x.+(\times(10^1)(1))(x) \rangle
    \qquad
    \langle \TT{thir}, \TT{:\!: num} \ \TT{-k}, 3 \rangle
    }
    {\langle \TT{teen}, \TT{: +k} \ \TT{num}, \lambda x.+(\times(10^1)(1))(x) \rangle
    \langle \TT{thir}, \TT{: -k}, 3 \rangle
    } \: \text{merge-3} \nonumber \\
&&\dfrac{\langle \TT{teen}, \TT{: +k} \ \TT{num}, \lambda x.+(\times(10^1)(1))(x) \rangle
    \langle \TT{thir}, \TT{: -k}, 3 \rangle
    }
    {\langle \TT{thirteen}, \TT{: num}, (\lambda x.+(\times(10^1)(1))(x))(3) \rangle \:,
    } \: \text{move-1} \:, \nonumber \\
\end{eqnarray}
while its ambiguous counterpart
\begin{eqnarray}\label{eq:deri13_4}
&&\dfrac{\langle \TT{teen}, \TT{: =num} \ \TT{+k} \ \TT{num}, \lambda x.+(\times(10^1)(1))(x) \rangle
    \qquad
    \langle \TT{three}, \TT{:\!: num}, 3 \rangle
    }
    {\langle \TT{threeteen},  \TT{: +k} \ \TT{num}, (\lambda x.+(\times(10^1)(1))(x))(3) \rangle
    } \: \text{merge-2} \nonumber \\
\end{eqnarray}
cannot be further processed due to a lacking licensee $\TT{-k}$.

The same argument applies to the ambiguous entries $\langle \TT{five}, \TT{:\!: num}, 5 \rangle $ and $\langle \TT{fif}, \TT{:\!: num} \ \TT{-k}, 5 \rangle$ where only the latter successfully derives $\langle \TT{fifteen}, \TT{: num}, +(\times(10^1)(1))(5)  \rangle$. Note that the currently learned grammar also derives the exponent $\TT{eightteen}$ instead of $\TT{eighteen}$; this could be corrected by either learning an additional entry $\langle \TT{eigh}, \TT{:\!: num} \ \TT{-k}, 8 \rangle$ and revising $\langle \TT{eight}, \TT{:\!: num}, 8 \rangle$, or, perhaps more appropriately, by introduction of additional phonotactical rules operating on abstract graphon representations \cite{WolffEichnerHoffmann01}. Moreover, since simplex numerals such as $\TT{four}$, $\TT{six}$, $\TT{seven}$, and $\TT{nine}$ must not possess any other features than $\TT{num}$, they would be doubled in a more rigorous treatment, resulting in four additional lexicon entries.

From a semantic point of view, the lexicon state in \Tab{tab:mg2} is not yet satisfactory, because another step of lambda abstraction can be applied to entry $\langle \TT{teen}, \TT{: =num} \ \TT{+k} \ \TT{num}, \lambda x.+(\times(10^1)(1))(x) \rangle$, entailing the semantics of plain addition
\begin{equation}\label{eq:addsem}
    \lambda x.+(\times(10^1)(1))(x) = (\lambda y.\lambda x.+(y)(x))(\times(10^1)(1)) \:.
\end{equation}

Incorporating this into the training process gives another updating dynamics
\begin{eqnarray}
    X_m &\gets& X_{m - 1} \setminus \{ \langle \TT{teen}, \TT{: =num} \ \TT{+k} \ \TT{num}, \lambda x.+(\times(10^1)(1))(x) \rangle \} \label{eq:dynm_1}  \\
    X_m &\gets& X_m \cup \{ \langle \varepsilon, \TT{:\!: =num} \ \TT{=num} \ \TT{+k} \ \TT{num}, \lambda y.\lambda x.+(y)(x) \rangle \} \label{eq:dynm_2} \\
    X_m &\gets& X_m \cup \{ \langle \TT{teen}, \TT{:\!: num}, \times(10^1)(1)) \rangle \} \label{eq:dynm_3} \:,
\end{eqnarray}
such that \pref{eq:dynm_1} removes the original $\TT{teen}$ from the lexicon which is subsequently replaced by the phonetically void addition operator $\langle \varepsilon, \TT{:\!: =num} \ \TT{=num} \ \TT{+k} \ \TT{num}, \lambda y.\lambda x.+(y)(x) \rangle$  and a new representative $\langle \TT{teen}, \TT{:\!: num}, \times(10^1)(1)) \rangle $.

Table \ref{tab:mg3} shows the updated lexicon at some even later time $t = m$.

\begin{table}[H]
\caption{\label{tab:mg3} Content of the minimalist lexicon $X_m$ of language learner $L$ after semantic reorganization at time $m$.}
\[
\begin{array}{lll}
    \langle \TT{one}, \TT{:\!: num}, 1 \rangle &
    \langle \TT{two}, \TT{:\!: num}, 2 \rangle &
    \langle \TT{three}, \TT{:\!: num}, 3 \rangle \\
    \langle \TT{thir}, \TT{:\!: num} \ \TT{-k}, 3 \rangle &
    \langle \TT{four}, \TT{:\!: num} \ \TT{-k}, 4 \rangle &
    \langle \TT{five}, \TT{:\!: num}, 5 \rangle \\
    \langle \TT{fif}, \TT{:\!: num} \ \TT{-k}, 5 \rangle &
    \langle \TT{six}, \TT{:\!: num} \ \TT{-k}, 6 \rangle &
    \langle \TT{seven}, \TT{:\!: num} \ \TT{-k}, 7 \rangle \\
    \langle \TT{eight}, \TT{:\!: num} \ \TT{-k}, 8 \rangle &
    \langle \TT{nine}, \TT{:\!: num}\ \TT{-k}, 9 \rangle &
    \langle \TT{ten}, \TT{:\!: num}, 10^1 \rangle \\
    \langle \TT{teen}, \TT{:\!: num}, \times(10^1)(1) \rangle &
    \langle \TT{eleven}, \TT{:\!: num}, +(\times(10^1)(1))(1) \rangle &
    \langle \TT{twelve}, \TT{:\!: num}, +(\times(10^1)(1))(2) \rangle \\
    \langle \varepsilon, \TT{:\!: =num} \ \TT{=num} \ \TT{+k} \ \TT{num}, \lambda y.\lambda x.+(y)(x) \rangle
\end{array}
\]
\end{table}

Now, the correct derivation of $\TT{thirteen}$ reads
\begin{eqnarray}\label{eq:deri13_5}
&&\dfrac{\langle \varepsilon, \TT{:\!: =num} \ \TT{=num} \ \TT{+k} \ \TT{num}, \lambda y.\lambda x.+(y)(x) \rangle
    \qquad
    \langle \TT{teen}, \TT{:\!: num}, \times(10^1)(1) \rangle
    }
    {\langle \TT{teen}, \TT{:} \TT{=num} \ \TT{+k} \ \TT{num}, (\lambda y.\lambda x.+(y)(x))(\times(10^1)(1)) \rangle
    } \: \text{merge-1} \nonumber \\
&&\dfrac{\langle \TT{teen}, \TT{:} \TT{=num} \ \TT{+k} \ \TT{num}, \lambda x.+(\times(10^1)(1))(x) \rangle
    \qquad
    \langle \TT{thir}, \TT{:\!: num} \ \TT{-k}, 3 \rangle
    }
    {\langle \TT{teen}, \TT{:} \TT{+k} \ \TT{num}, \lambda x.+(\times(10^1)(1))(x) \rangle
    \langle \TT{thir}, \TT{: -k}, 3 \rangle
    } \: \text{merge-3} \nonumber \\
&&\dfrac{\langle \TT{teen}, \TT{:} \TT{+k} \ \TT{num}, \lambda x.+(\times(10^1)(1))(x) \rangle
    \langle \TT{thir}, \TT{: -k}, 3 \rangle
    }
    {\langle \TT{thirteen}, \TT{:} \TT{num}, +(\times(10^1)(1))(3) \rangle \:.
    } \: \text{move-1} \:. \nonumber \\
\end{eqnarray}

By virtue of lexicon $X_m$ the learner is able to correctly reproduce numerals $\TT{one}, \dots, \TT{nineteen}$, employing its UMT \cite{GrabenMeyerEA19}. This will be rewarded by the teacher. Later, the teacher utters the UMPs $u_{20} = \langle \TT{twenty}, \times(10^1)(2) \rangle$, $u_{21} = \langle \TT{twentyone}, +(\times(10^1)(2))(1) \rangle$, $u_{22} = \langle \TT{twentytwo}, +(\times(10^1)(2))(2) \rangle$ etc. Again, the learner will first incorporate $\langle \TT{twenty}, \times(10^1)(2) \rangle$ according to rule \pref{eq:dyn1} into the lexicon. But upon perceiving $u_{21}$ its pattern matching device produces a common morpheme
\begin{equation}\label{eq:ty}
      \langle \TT{ty}, \TT{:\!:} \: \TT{=num} \: \TT{num}, \lambda x. \times(10^1)(x) \rangle
\end{equation}
through lambda abstraction. Then the essentially same processes of reinforcement learning are repeated as above until the complete numeral system of the language taught by the teacher has been acquired by the learner.


\section{Discussion}
\label{sec:disc}

In this contribution we have outlined an algorithm for effectively learning the syntactic morphology and semantics of English numerals \cite{Hurford01}. Number words are presented to a cognitive agent by a teacher in form of utterance meaning pairs (UMP) where the meanings are encoded as arithmetic terms from a suitable term algebra. This representation allows for the application of compositional semantics via lambda calculus. For the description of syntactic categories we use Stabler's minimalist grammar (MG) \cite{Stabler97, StablerKeenan03}, a powerful computational implementation of Chomsky's recent Minimalist Program for generative linguistics \cite{Chomsky95}. Despite the controversy between Chomsky and Skinner \cite{Chomsky59}, we exploit reinforcement learning \cite{Skinner15, SuttonBarto18} as training paradigm. Since MG encodes universal linguistic competence through the five inference rules (\ref{merge-1} -- \ref{move-2}), thereby separating innate linguistic knowledge from the contingently acquired lexicon, our approach could potentially unify generative grammar and reinforcement learning, hence resolving the abovementioned dispute.

Minimalist grammar can be learned from linguistic dependency structures \cite{KobeleCollierEA02, StablerEA03, BostonHaleKuhlmann10, KleinManning04} by positive examples, which is supported by psycholinguistic findings on early human language acquisition \cite{Ellis06, Pinker95, Tomasello06}. However, as Pinker \cite{Pinker95} has emphasized, learning through positive examples alone, could lead to undesired overgeneralization. Therefore, reinforcement learning that might play a role in children language acquisition as well \cite{Moerk83, SundbergEA96}, could effectively avoid such problems. The required dependency structures are directly provided by the semantics in the training UMPs. Thus, our approach is explicitly semantic-driven, in contrast to the algorithm in \cite{KleinManning04} that regards dependencies as latent variables for EM training.

As a proof-of-concept we suggested an algorithm for English numerals. However, we also have evidence that it works for German and French number systems as well and hopefully for other languages also. Using attribute-value logics \cite{Johnson88} and its associated term algebra, it should be possible to encode the semantics of arbitrary utterances in a compositional fashion. This will open up an entirely new avenue for the further development of speech-controlled cognitive user interfaces \cite{TschopeDuckhornEA18}.




\end{document}